# REAL-TIME SEAFLOOR SEGMENTATION AND MAPPING

Michele Grimaldi[1,3], Nouf Alkaabi[1,2], Francesco Ruscio[4,5], Sebastian Realpe Rua[1], Rafael Garcia[1], Nuno Gracias[1]

[1]Computer Vision and Robotics Research Institute (VICOROB), University of Girona, 17003 Girona, Spain
[2]Autonomous Robotics Research Center, Technology Innovation Institute (TII), Abu Dhabi, United Arab Emirates
[3]School of Engineering and Physical Sciences, Heriot-Watt University, EH144AS Edinburgh, UK
[4]Dipartimento di Ingegneria dell'Informazione (DII), University of Pisa, via Girolamo Caruso 16, 56122, Pisa, Italy
[5]Interuniversity Center of Integrated Systems for the Marine Environment (ISME), Italy

**ABSTRACT:**

Posidonia oceanica meadows are a species of seagrass highly dependent on rocks for their survival and conservation. In recent years, there has been a concerning global decline in this species, emphasizing the critical need for efficient monitoring and assessment tools. While deep learning-based semantic segmentation and visual automated monitoring systems have shown promise in a variety of applications, their performance in underwater environments remains challenging due to complex water conditions and limited datasets. This paper introduces a framework that combines machine learning and computer vision techniques to enable an autonomous underwater vehicle (AUV) to inspect the boundaries of Posidonia oceanica meadows autonomously. The framework incorporates an image segmentation module using an existing Mask R-CNN model and a strategy for Posidonia oceanica meadow boundary tracking. Furthermore, a new class dedicated to rocks is introduced to enhance the existing model, aiming to contribute to a comprehensive monitoring approach and provide a deeper understanding of the intricate interactions between the meadow and its surrounding environment. The image segmentation model is validated using real underwater images, while the overall inspection framework is evaluated in a realistic simulation environment, replicating actual monitoring scenarios with real underwater images. The results demonstrate that the proposed framework enables the AUV to autonomously accomplish the main tasks of underwater inspection and segmentation of rocks. Consequently, this work holds significant potential for the conservation and protection of marine environments, providing valuable insights into the status of Posidonia oceanica meadows and supporting targeted preservation efforts

## 1. INTRODUCTION

In recent years, significant advancements have been made in the field of Autonomous Underwater Vehicles (AUVs). These technological improvements have expanded visual data collection capabilities, enabling researchers and experts to access larger and more challenging underwater environments and obtain real images. The enhanced capabilities of AUVs have revolutionized the way underwater exploration and data collection are conducted. These vehicles have advanced imaging systems and sensors that can capture high-resolution images and collect valuable data from previously inaccessible underwater environments. By navigating through diverse and challenging terrains, AUVs provide researchers with a unique opportunity to explore and document underwater ecosystems in unprecedented detail (Williams, 2012).

The advancements in visual data collection have significant implications for various fields of study. Environmental researchers can monitor and analyze changes in underwater environments, gaining valuable insights into the intricate and hidden world of the seabed with improved accuracy and efficiency (Martin-Abadal et al., 2018a). Additionally, these advancements contribute to a better understanding of underwater ecosystems. They enhance efficiency in research and conservation efforts by providing a valuable understanding of the distribution and behaviour of marine species, such as Posidonia oceanica, and the overall health of the marine ecosystem. Furthermore, this technology supports the sustainable management of our marine ecosystem by aiding in effective decision-making and conservation strategies.

Regrettably, reports indicate that the total area of meadows has decreased by 34% over the past 50 years. This indicates that the decline is a widespread issue primarily caused by the cumulative impacts of various local stressors (Telesca et al., 2015). The findings emphasise the significance of conducting surveys to evaluate the current condition and prioritise areas for implementing effective strategies to reduce threats. These measures aim to reverse the current trends and ensure the long-term survival of Posidonia oceanica across the Mediterranean region (Barcelona et al., 2021)(Serra et al., 2020). Given its ecological importance, it is necessary to comprehend and conserve it and its habitat. Understanding and preserving this seagrass species has a huge weight in maintaining healthy coastal ecosystems and preserving biodiversity.

Recently, (Ruscio et al., 2023) explored the use of the Mask R-CNN model for segmenting Posidonia meadows, demonstrating its effectiveness in segmentation tasks. Furthermore, image segmentation enabled an Autonomous Underwater Vehicle (AUV) to implement a tracking strategy that generated guidance references for tracking the contours of seagrass, serving as a valuable tool for marine environment conservation. Building upon the latter research, in this paper we introduce the inspection framework to monitor and map seabeds containing Posidonia seagrass.

## 2. RELATED WORK

The segmentation and detection of Posidonia oceanica in underwater environments have seen significant advancements, driven

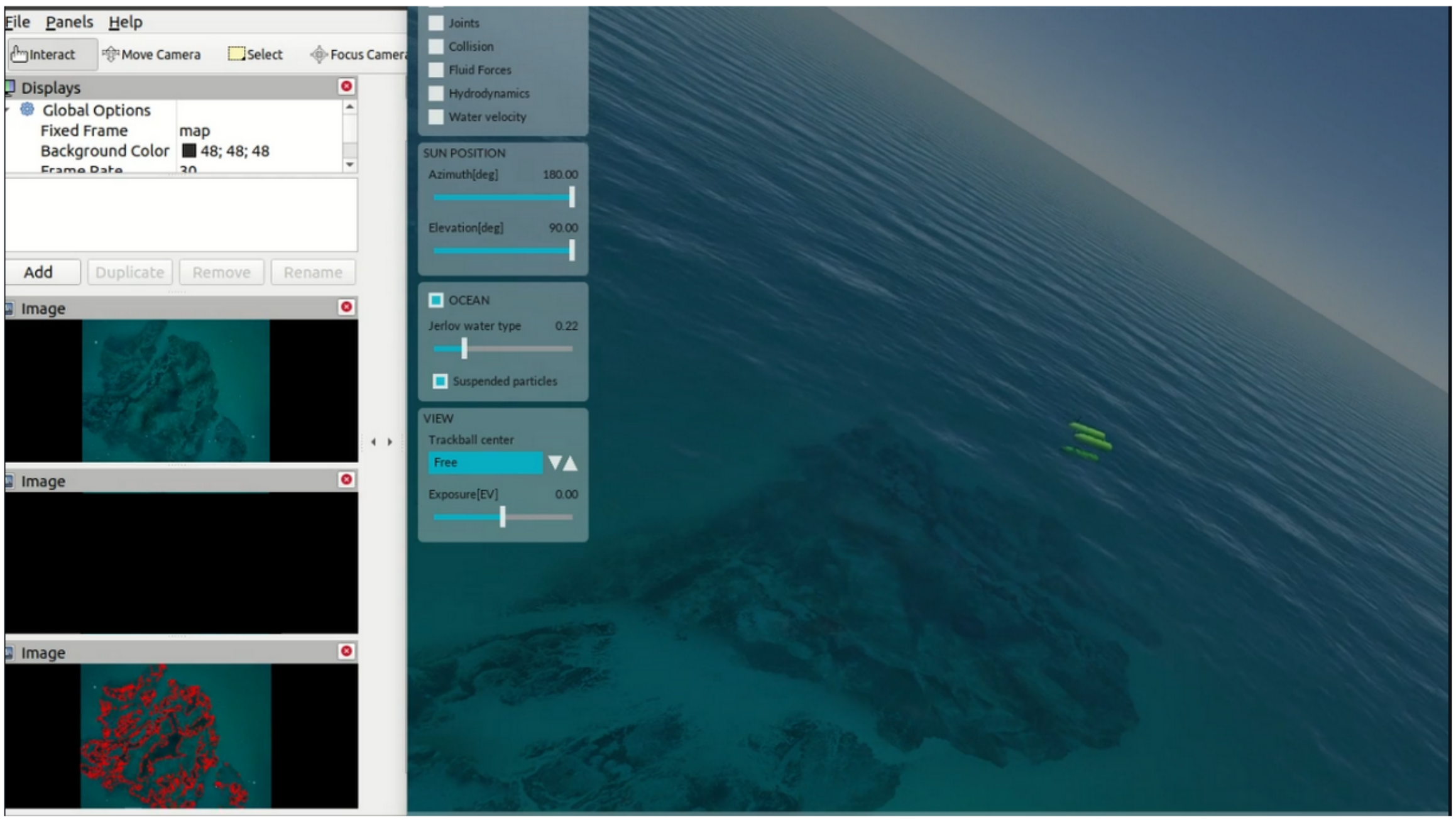

Figure 1. Girona500 detecting a dark patch on a rock site. The robot is going to dive to check if there is a presence of Posidonia meadows. On the left side, there are the images from the robot's camera (top), and the image with the detected dark patches (bottom). The middle window will show the segmented Posidonia meadow is detected

by the need for accurate mapping and monitoring of this vital seagrass species. As a fundamental biological indicator, P. oceanica meadows are very important for assessing the health of marine ecosystems and play an essential role in conserving coastal morphology. The composition, extent, and structure of these meadows are influenced by various environmental factors, including substrate type, seabed geomorphology, hydrodynamics, depth, light availability, and sedimentation rates. Consequently, effective monitoring of these meadows is critical for environmental conservation efforts.

Recent advancements have introduced a variety of methodologies aimed at enhancing the efficiency and accuracy of P. oceanica monitoring, reducing the need for traditional, labour-intensive, and potentially hazardous data collection methods involving scuba divers. One approach leverages highly autonomous marine robotic platforms, such as Remotely Operated Vehicles (ROVs), equipped with multi-parametric sensors to automatically detect and map underwater vegetation, including this seagrass. This method, which combines acoustic data with video imagery, enables the reconstruction of a 2.5D model of the sea bottom, facilitating detailed environmental monitoring. (Ferretti et al., 2017)

Another significant development involves the application of deep neural networks for the semantic segmentation of P. oceanica and its seafloor habitat (Martin-Abadal et al., 2019, Martin-Abadal et al., 2018b). Utilizing the VGG16 CNN (Simonyan and Zisserman, 2014) for feature extraction and the FCN8 architecture (Long et al., 2015) for generating confidence maps, this method has shown high accuracy in segmenting the seagrass meadows, demonstrating the robustness of CNNs (LeCun et al., 2015) in underwater image analysis.

In (Gonzalez-Cid et al., 2021) CNNs are used to explore large areas colonized by this seagrass. By creating 2D and 3D maps and automatically quantifying the bottom coverage of P. oceanica, this method significantly reduces costs, infrastructure needs, and safety risks compared to traditional diver-based monitoring. It also enhances the mission scope, data quality, and processing accuracy, representing a new paradigm in benthic marine habitat assessment. Furthermore, (Burguera, 2020) has introduced two models: Model I, a CNN, and Model II, a simplified version without convolutional layers. These models, combined with a region-growing algorithm, have shown superior performance in detecting P. oceanica, particularly in coastal areas of Mallorca, offering higher detection rates and faster processing speeds than previous methods.

In (Bonin-Font et al., 2017) integration of Autonomous Underwater Vehicles (AUVs) with bottom-looking cameras has been explored to advance the seagrass monitoring. This technique involves training classifiers with Gabor filter image patch descriptors and support vector machines (SVMs) to autonomously detect P. oceanica in individual images. The process is enhanced by photo-mosaicking to create comprehensive coverage maps, achieving high precision in detecting various seagrass patterns across different environmental conditions.

In (Ruscio et al., 2021), a Smart Dive Scooter (Costanzi et al., 2019) equipped with visual acquisition and acoustic localization systems was used. An acoustic-based geo-referencing strategy was employed, leveraging the synchronization between the camera's audio track and transponder pings for acoustic positioning. The resulting diver path was refined using visual odometry and further improved with a Rauch-Tung-Striebel smoother. The processed data produced an image mosaic and a qualitative distribution map of the seagrass, demonstrating the effectiveness of this process in supporting divers during monitoring activities.

Finally, a photogrammetric approach has been proposed for the rapid and reliable characterization of the seabed, with particular reference to P. oceanica coverage (Russo et al., 2023). This methodology aims to reduce the effects of environmental factors on underwater images, such as the bluish or greenish tints typical of underwater environments (Grimaldi et al., 2023). By applying two different algorithms to restore the images, the resulting 3D point cloud allowed for better categorization of the surveyed area compared to traditional image processing techniques. This enhanced workflow provides a more effective means of monitoring and mapping P. oceanica meadows, contributing to a deeper understanding of the seagrass's spatial distribution and health.

## 3. DATASETS

A diverse dataset is essential for accurately classifying underwater species and optimizing deep-learning models for underwater object recognition. It ensures that the deep-learning model recognizes a wide variety of distinct features and appearances, improving its capacity to detect and classify objects accurately. Additionally, the model becomes more resilient to changes in appearance, lighting, and other environmental factors, enabling it to perform effectively in different underwater settings and adapt to the challenges posed by the complexity and dynamic nature of the marine environment. For this reason, the dataset was acquired at distinct locations and altitudes above the seafloor, using different cameras and under varying environmental conditions, such as visibility and illumination. Of the dataset previously obtained, 86% primarily consisted of images of Posidonia oceanica and debris from (Ruscio et al., 2023). These images were captured using a waterproof GoPro Hero 2 Action Cam (1920x1080 pixels) mounted on an AUV at various altitudes along the coast of the Croatian island of Murter. Additionally, during the monitoring of meadows in the coastal region of Rapallo, Italy, some images were taken by a diver using four GoPro Hero 5 cameras, each with a resolution of 2704x1520 pixels, mounted on a scooter. Others were recorded using the bottom-looking camera (640x505 pixels) of the Zeno AUV. The remaining images were captured in Costa Brava, Spain, using the onboard camera of the Sparus II AUV, which has a resolution of 1936x1464 pixels.

The remaining 14% of the dataset is dedicated to the class of rocks. These images, with a resolution of 3.2 megapixels, were captured by divergent stereo cameras attached to the Girona1000 AUV. They were collected in October 2022 along the Catalan coast near San Feliu de Guixols, where the AUV navigated approximately 4 to 6 meters above the seafloor. Other images were collected in very shallow waters of up to 3 meters depth using a GoPro Hero 2 camera with a resolution of 1920x1080 pixels. These were obtained in Palamos on the Catalan coast and in Guadeloupe in the French Antilles in 2017. Table 1 provides an overview of the dataset's characteristics.

### 3.1 Image Enhancement

Some images posed challenges due to the distortion caused by light reflections on the seabed and the inherent blurriness induced by the water medium. Besides, the presence of rocks covered in sand and plants resembling debris adds complexity. These challenges have the potential to create confusion during the model's training process, thereby affecting the accuracy of rock recognition.

To address these challenges, the diverse rock dataset was preprocessed using histogram equalization and inverse gamma correction. By reducing the effect of light reflections and enhancing the visibility of rocks against the background, histogram equalization improved the contrast of the underwater images. Inverse gamma correction was used to correct image non-linearities, enhancing the features of rocks, and assisting in their precise detection. The combined application of these techniques aimed to optimize the underwater images, allowing the model to capture important features that might have been obscured or difficult to discern in the original images. As a result, the model can acquire more precise and robust representations of rocks despite the challenges, which facilitates accurate model learning.

### 3.2 Data Labeling

The images of rocks and Posidonia oceanica were accurately labelled using the open-source software makesense.ai. The annotations were saved in a JSON file, defining the precise boundaries of the Posidonia oceanica and rock regions within each image. To prepare the data for Mask R-CNN, the annotations were converted into binary masks. This mask format provides the necessary input for Mask R-CNN to perform instance segmentation accurately. Each pixel in the image was assigned a value of 0, 1, 2, or 3, representing whether it belongs to the background, Posidonia oceanica, debris, or rocks, respectively.

## 4. SEGMENTATION: ARCHITECTURE, TRAINING AND EVALUATION

### 4.1 Network Training

The dataset used for the 3 classes, Posidonia oceanica, Debris, and Rocks with a total of 6949 images, with binary mask labels used as ground truth during training. The entire dataset was randomly divided into three subsets: 70% training (4865 images), 20% validation (1389 images), and 10% testing (695 images). Then, the network training process followed a three-step approach: In the first session, starting from epoch 0 and continuing until epoch 440, the training commenced by leveraging pre-trained weights from the COCO detection dataset specifically for the rock images to test the model's capability to learn on these images and make accurate predictions. The models showed good prediction results on the test set. Subsequently, in the second session, the training was repeated using pre-trained weights from the COCO detection dataset. This time training encompassed the combined dataset, which included three classes: Posidonia oceanica, Debris, and Rocks, also starting from epoch 0 to 440. Unfortunately, due to dataset imbalance, the model couldn't predict the rocks in the image. The training progressed to the third session where the weights obtained from (Ruscio et al., 2023) were utilized. Lastly, the focus was on adding rocks as a new class to the existing model. In this case, the model showed an undesirable behaviour, resulting in false positives, where it incorrectly segments rocks as debris. This misclassification occurs because of the similarities between the debris colour and the sand covering the rocks. Table 3 shows the adopted hyper-parameters during the training of the Mask R-CNN model while table 2 depicts the different training sessions. In the fourth training session, we employed the pre-trained weights from (Ruscio et al., 2023) and trained the model exclusively on the enhanced rock images. In the fifth training session, we tackled the network imbalance caused by the limited number of rock images by applying geometric data augmentation techniques such as rotation, flipping, scaling, and zooming to the enhanced images. This process yielded a total of 10,822 training images, encompassing rocks, Posidonia oceanica, and debris.

Table 1. Characteristics of the dataset used to train the Mask R-CNN model

| Location | Camera | Image Size (Pixels) | Number of Images | Classes |
|---|---|---|---|---|
| Island of Murter, Croatia | GoPro Hero 2 Action Cam | 1920×1080 | 4190 | Posidonia, Debris |
| Rapallo, Italy | GoPro Hero 5 | 2704×1520 | 460 | Posidonia, Debris |
| Rapallo, Italy | Basler ace | 640×505 | 930 | Posidonia, Debris |
| Costa Brava, Spain | Teledyne Flir | 1936×1464 | 420 | Posidonia, Debris |
| Catalan coast | Stereo cameras | 3.2Mpixels | 66 | Rocks |
| Guadeloupe, French Antilles | GoPro Hero 2 Action Cam | 1920×1080 | 455 | Rocks |
| Palamós, Catalan Coast | GoPro Hero 2 Action Cam | 1920×1080 | 455 | Rocks |

Table 2. Different Network Training Sessions

| Training | Class Name | Number of Images | Number of Epochs | Best Epoch |
|---|---|---|---|---|
| Training 1 | Rocks | 976 | 0-440 | 256 |
| Training 2 | Posidonia, Debris, Rocks | 6949 | 0-440 | 268 |
| Training 3 | Posidonia, Debris, Rocks | 6949 | 440-600 | 511 |
| Training 4 | Rocks (image enhancement) | 976 | 440-660 | 509 |
| Training 5 | Posidonia, Debris, Rocks (image enhancement) | 10822 | 440-660 | 509 |

Furthermore, an additional trial of continue-training sessions was conducted on the enhanced rock images. It exhibited interesting results by classifying most of the rocks in the image as shown in 2 $d$). Even though some batches displayed peaks indicating challenging images and false positives, these images were not trained enough to achieve high performance. Overall, the model showed promising behaviour.

| Hyper-parameter | Value |
|---|---|
| Number of Classes | 3+1 (background) |
| RPN anchor scales | [32, 64, 128, 256, 512] |
| RPN anchor ratios | [0.5, 1, 2] |
| Optimiser | SGD (with momentum) |
| Momentum | 0.9 |
| Weight decay | 1.e−4 |
| Learning rate | 1.e−4 |
| Epochs | 110 |
| Batch size | 2 |

Table 3. Hyperparameters of the Pre-trained Network

### 4.2 Network Evaluation

The segmentation performance of the Mask R-CNN model was assessed by evaluating its accuracy. The evaluation used a measure called Intersection over Union (IoU), which compares the overlap between two regions. The IoU ratio, defined in Equation (1), compares the model's predicted mask to the ground truth labelled mask. Higher IoU values indicate better overlap.

$$IoU = \frac{groundtruth \cap prediction}{groundtruth \cup prediction} \tag{1}$$

A mean value of 0.824 was computed from the testing images set. This result is compelling evidence for the effective performance of the suggested classification model, particularly considering that the model had not previously encountered these specific images. Figures 2 and 3 show the results of the different training sessions.

## 5. INSPECTION FRAMEWORK

The inspection framework, depicted in figure 6 is composed of three nodes which will define the behaviour of the robot.

### 5.1 PathPlanner Node

This node creates the inspection trajectory 5 for the robot to follow during the inspection process, enabling it to navigate across the water's surface while ensuring comprehensive coverage of the environment.

### 5.2 Dark Patch Detection Node

This node works upon the detection of a region with reduced luminosity, indicative of a dark patch. To add some complexity and try to simulate real-world conditions, the input image is pre-processed using the Beer-Lambert law (Stavn, 1988), which allows to further simulate backscattering and challenging visibility conditions. Then the node will apply a first threshold to eliminate the white particles of the backscattering, then the image is converted in the HSV space (hue, saturation, value) and finally a threshold on the Value channel will allow us to find the contours of the black patches. The threshold and the Beer-Lambert law parameter are proportional to the actual depth of the robot. Furthermore, to eliminate the dark patch that is created by the reflection of the robot on the seabed, the dark patches that are in a bounding box with the centre equal to the centre of the image are not considered. Allowing the robot to descend towards the seabed near the target area to perform a segmentation process. It is worth noting that the node will create an ”alpha shape” using the points on the surface and the point at the bottom projected to the surface, to create a map of the “already explored areas.” This will prevent the robot from descending twice in the same area.

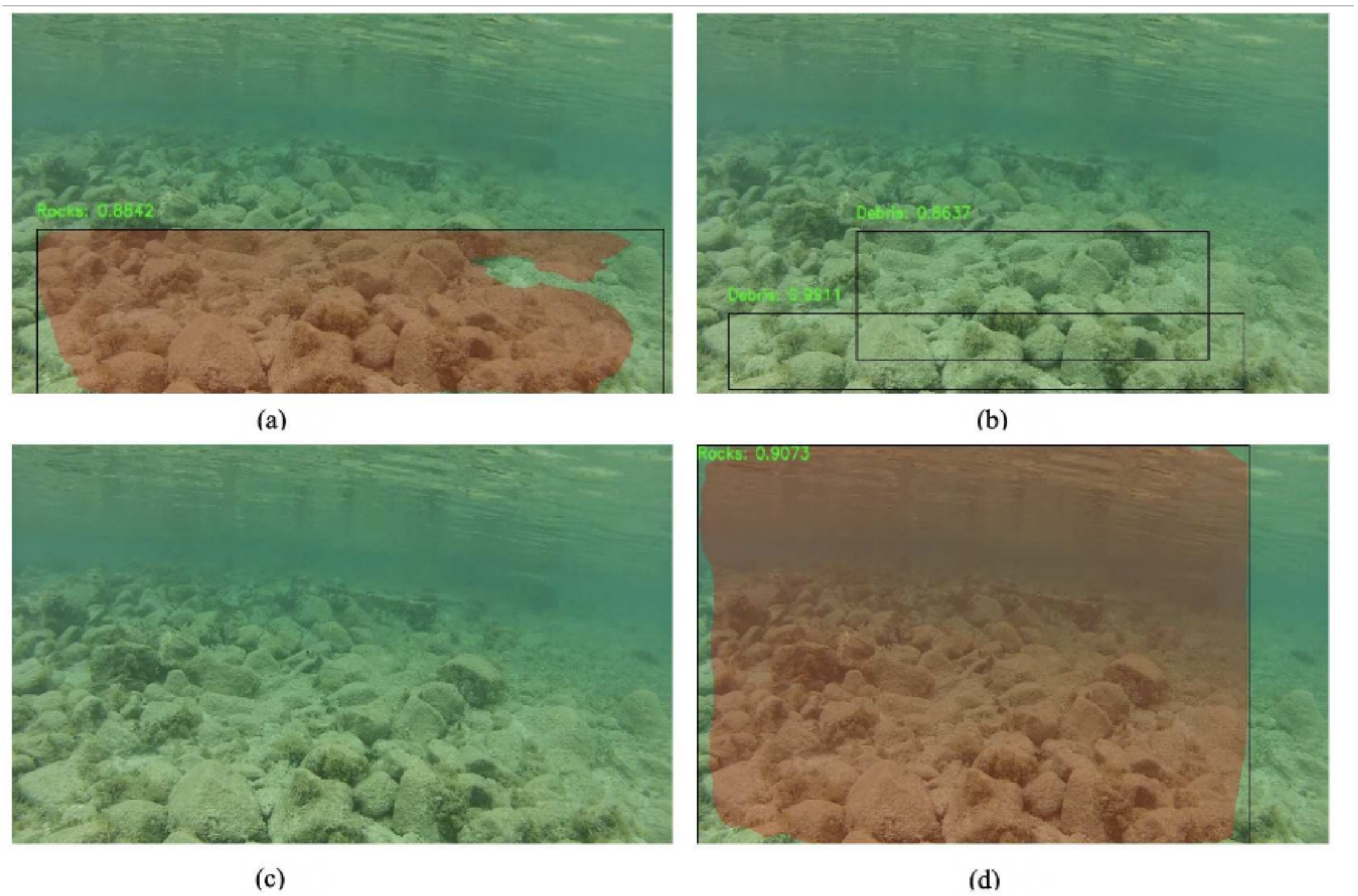

Figure 2. A comparative analysis highlighting the issues in each training session. (a) Good prediction in training session 1. (b) False positives in training session 2. (c) The model couldn't predict on the same image in training session 3. (d) Promising results on enhanced images.

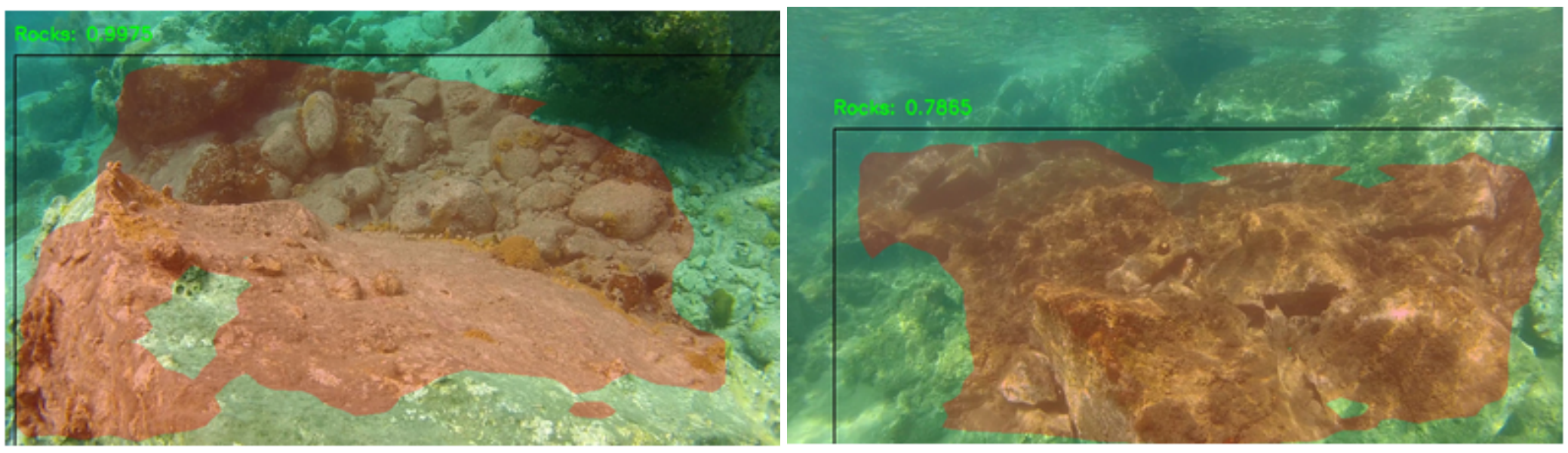

Figure 3. Results of the fifth training session demonstrated that rocks could be detected even when partially covered by sand or illuminated by sunlight patterns.

### 5.3 MaskRCNN-boundaries Node

This node takes input images and uses the Mask R-CNN model to perform image segmentation allowing for accurate detection, the model will analyse the segmented data, enabling the identification of Posidonia oceanica meadows. In the case of Posidonia oceanica detection, the node activates a contouring procedure specifically tailored for these underwater meadows generating reference values for the control system. The robot depth can be set and in our experiment was set to 5m from the seabed which is at 15m from the sea level as shown in 4.

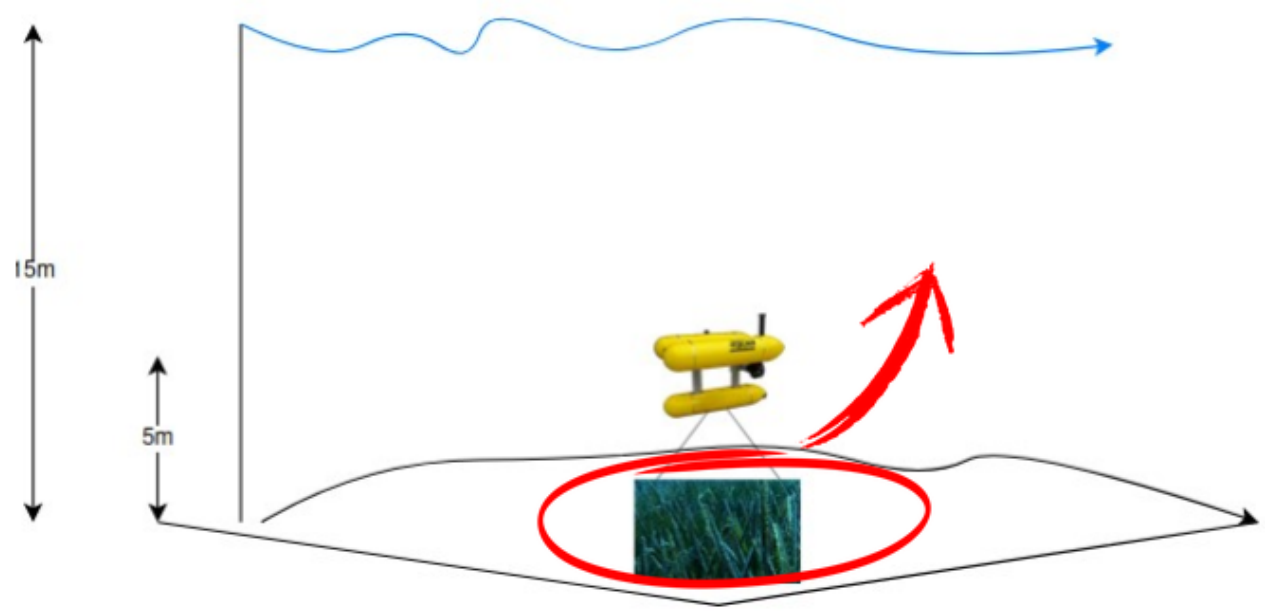

Figure 4. Contour following

## 6. EXPERIMENT

### 6.1 Simulation Environment and Setup

The open-source Stonefish C++ simulation library (Cieślak, 2019),(Grimaldi et al., 2025) combined with the interface package Stonefish_ros Robot Operating System (ROS) were used in this work to validate the inspection framework. Stonefish is specially designed for marine and surface robots. It supports buoyancy, collision detection, geometry-based hydrodynamics, and rigid body dynamics. To replace the actual system with a simulated robot, it simulates all underwater sensors and actuators as well. Additionally, the library incorporates light absorp-tion and scattering models, enabling the software to generate authentic underwater images, and enhancing the fidelity of the simulation. The main advantage of using Stonefish is to closely mimic the complexity and realism of an underwater environment.

The Girona 500 AUV (Ribas et al., 2012) is used for the simulation task, specifically for intervention tasks and monitoring surveys. To ensure an accurate simulation of the vehicle's physical characteristics and forces, the simulator incorporated the geometric model and material properties of its components. The simulator also integrated the complete software architecture of the AUV, which is based on the Robot Operating System (ROS) ensuring integration between the proposed framework and the actual system of the vehicle.

During the simulations, the AUV configuration consisted of five thrusters providing control over all degrees of freedom except for roll motion. The stability in the roll was achieved through careful weight distribution and flotation devices. For navigation, the AUV was equipped with a Doppler Velocity Log (DVL), an Inertial Measurement Unit (IMU), and a pressure sensor. These sensors' measurements were fused using an Extended Kalman Filter (EKF)-based navigation algorithm to accurately estimate the AUV's motion. Additionally, the AUV was equipped with a bottom-looking camera situated in the lower hull to capture images of the seafloor.

A virtual environment was created to evaluate the performance of the Mask R-CNN model, which was an extension of the work conducted in (Ruscio et al., 2023), who developed two distinct mosaics as part of his study. These mosaics were specifically designed to simulate varying distributions of seagrass on the seafloor. However, to thoroughly test the detection capabilities of the Mask R-CNN model, the virtual environment was expanded to four times its original size. Within this enlarged virtual environment, five isolated patches were strategically placed at distinct locations, allowing the robot to thoroughly explore and interact with the environment. These patches were designed to represent different scenarios: some consisted solely of rocks, others solely of Posidonia oceanica, and some contained a combination of rocks and Posidonia oceanica. By incorporating these variations, the virtual environment aimed to closely mimic the complexity and realism of an underwater environment.

A mission was then set creating a reference point using the software IQUA View [1]. These reference points guide the Autonomous Underwater Vehicle (AUV) along the desired trajectory, within the identified meadow. Conversely, if the model identifies the presence of rocks, the robot will ascend and continue its trajectory toward the subsequent waypoints while adhering to the procedure. This iterative process continues until the completion of the predefined trajectory.

## 7. RESULTS

The experiment begins with a predefined trajectory as shown in 5, featuring a specific orientation and speed. The vehicle initiates its inspection along this trajectory, actively searching for dark patches. Upon detecting a dark patch, it descends closer to the seabed, enabling the model to segment these patches and determine the presence of Posidonia meadows. Once detected, the vehicle initiates a boundary-tracking procedure. Otherwise, in the case of rocks, the AUV elevates and continues inspection until the entire trajectory path has been covered.

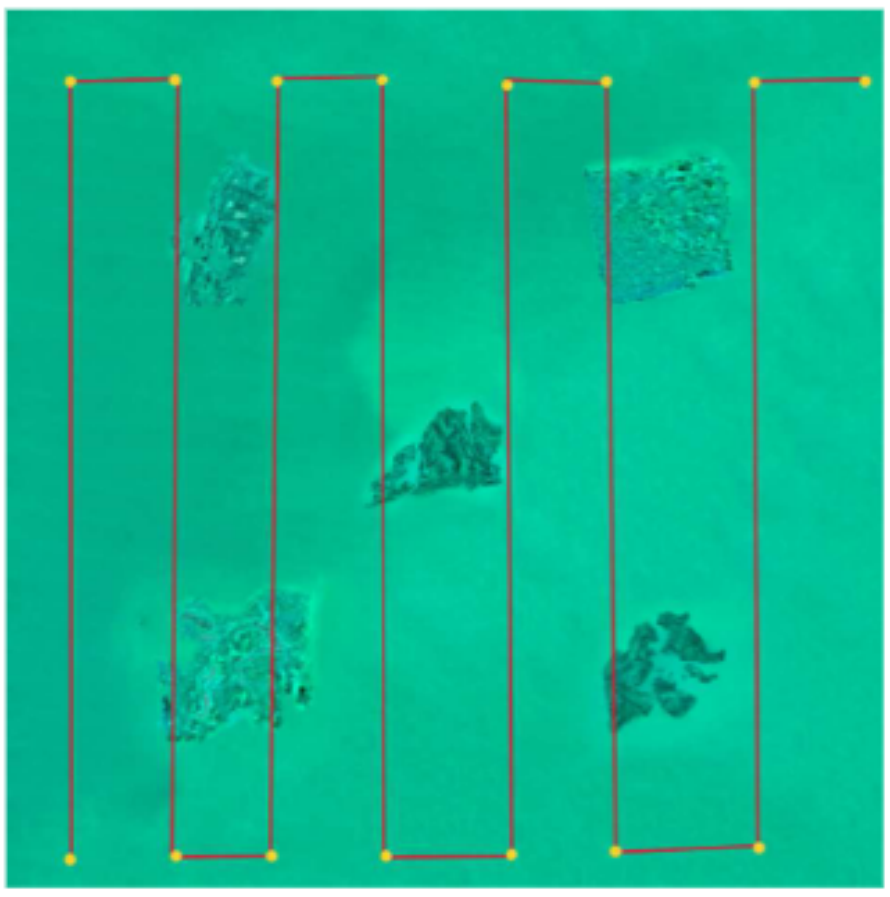

Figure 5. The inspection trajectory

Two outcomes were observed in the experiment. The first outcome is that the model successfully identified dark patches as expected, which means that the AUV can dive and arise depending on what the dark patch is composed of. The second one is that the network was not able at first to detect the rocks and Posidonia meadows. This is because Stonefish simulates Jerlov water types (Jerlov, 1968). These water types can be defined either by their irradiance transmission through a given thickness of water, or via the diffuse attenuation coefficient averaged over a given depth. Each water type has its properties, which significantly impact the image perception in the presence of water and its interaction with light. As a result, the presence of water and its interaction with light significantly impact the image perception of rocks. These distinct attributes introduce variations in the image appearance that differ from the training data initially provided to the model. Consequently, the model encounters difficulty in accurately detecting rocks and meadows within the simulated environment. We added the images from the simulated Stonefish environment to the model's training dataset to solve this issue.

## 8. CONCLUSION

This study introduced an inspection framework that could be used to monitor seabeds, search for areas containing Posidonia meadows, and map them. The dark patches detector is used to avoid areas containing only rocks and only map areas that are meaningful to monitor. The framework integrates an improved Mask R-CNN model with an autonomous underwater vehicle (AUV) including a new class for rock segmentation. A video showing the performance of the tool can be found at this address: `https://drive.google.com/file/d/1FRyNY8swtXrbUQYesDI5HEACoHtEnHyZ/view?usp=sharing`.
This approach provides a reliable tool for continuous monitoring and offers valuable insights to support conservation efforts, ensuring the protection of Posidonia oceanica meadows for the future.

## 9. ACKNOWLEDGEMENT

The Spanish government supported this work through the SIREC "Seafloor Intelligent Robot Exploration and Classific-

[1] https://iquarobotics.com/iquaview-graphical-user-interface

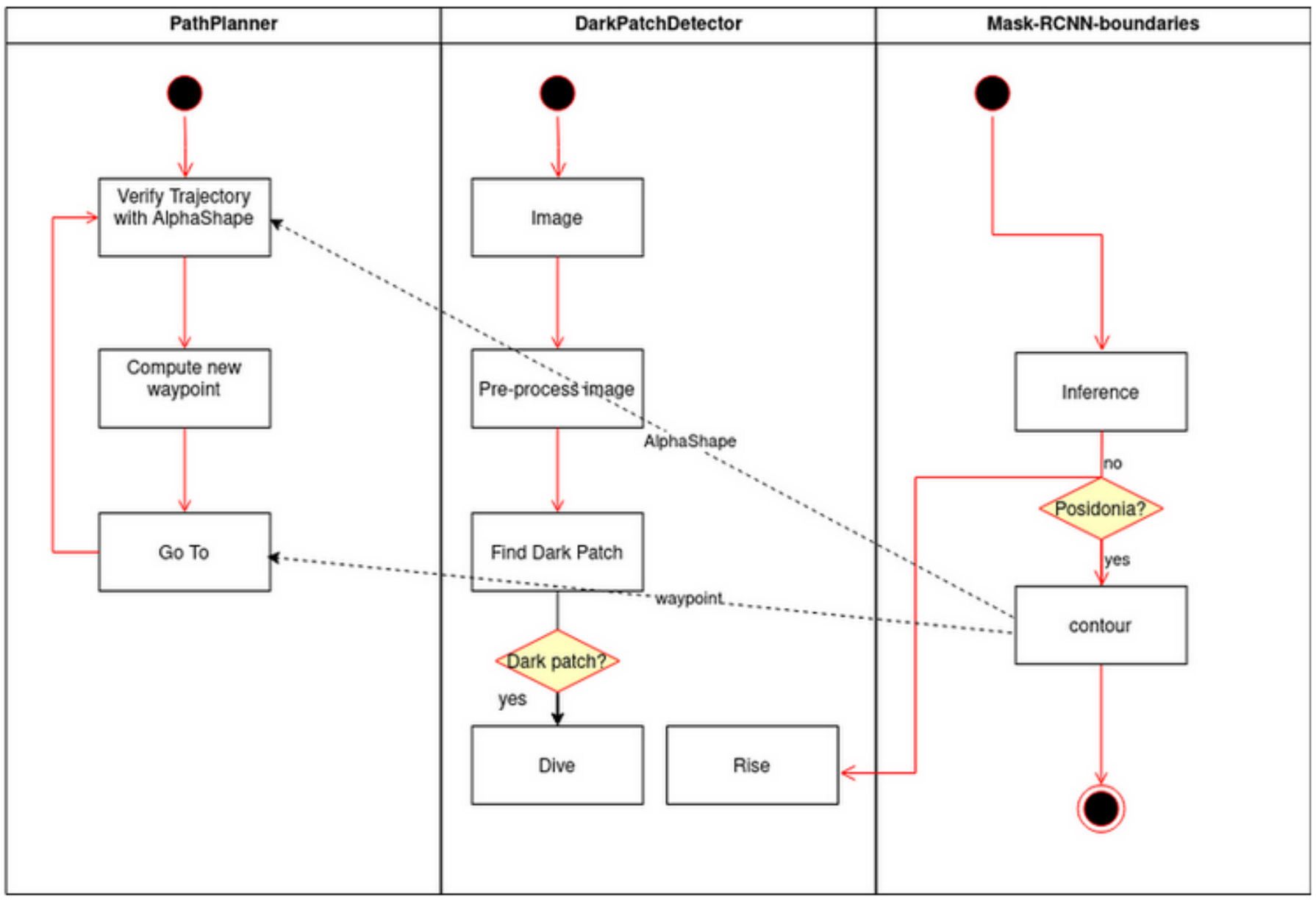

Figure 6. Nodes composing the inspection framework

ation" Project with reference: PID2020-116736RB-IOO. It has also been partially supported by the project "IURBI - Intelligent Underwater Robot for Blue Carbon Inventorying" (Ref. CNS2023-144688), funded by the Spanish Ministerio de Ciencia, Innovación y Universidades.